\documentclass{article} 
\usepackage{iclr2027_conference,times}

\iclrfinalcopy

\usepackage{amsmath,amsfonts,bm}

\def\eqref#1{equation~\ref{#1}}
\def\Eqref#1{Equation~\ref{#1}}

\def\1{\bm{1}}

\DeclareMathAlphabet{\mathsfit}{\encodingdefault}{\sfdefault}{m}{sl}
\SetMathAlphabet{\mathsfit}{bold}{\encodingdefault}{\sfdefault}{bx}{n}

\usepackage{amsmath}
\usepackage{hyperref}
\usepackage{url}
\usepackage{booktabs}
\usepackage{multirow}
\usepackage{xcolor}
\usepackage{colortbl}
\usepackage{graphicx}
\usepackage{wrapfig}
\usepackage{needspace}
\usepackage{amsmath,amssymb}
\usepackage{etoolbox}
\usepackage{csquotes}
\AtBeginEnvironment{table}{\setlength{\abovecaptionskip}{0pt}\setlength{\belowcaptionskip}{6pt}}
\AtBeginEnvironment{table*}{\setlength{\abovecaptionskip}{0pt}\setlength{\belowcaptionskip}{6pt}}

\newcommand{\ours}{HOPD}

\title{Hesitation-Aware On-Policy Distillation for Diffusion Language Models}

\author{
Jianguo Huang$^{1}$,
Lipeng Wan$^{2}$,
Yanchen Deng$^{1}$\thanks{Corresponding author},
Bo An$^{1}$ \\
$^{1}$Nanyang Technological University, Singapore
$^{2}$Xi'an Jiaotong University, China
}

\begin{document}

\maketitle

\begin{abstract}

Diffusion large language models (dLLMs) generate text by iterative unmasking. At each denoising step, a dLLM proposes a token at every masked position, but the decoder commits only a confident subset of these proposals. Trace-based on-policy distillation (TOPD) builds on this process by matching the student to a stronger teacher, yet only at the committed positions. We argue that this discards much of the useful signal, which resides in the uncommitted proposals, where the student has made a prediction but is not yet confident enough to commit it. We call these proposals hesitations. In our pilot study on an SDAR-4B student, hesitations make up only 24\% of supervisable state–position pairs but carry 66\% of the teacher–student divergence. To exploit this signal, we propose \textit{Hesitation-Aware On-Policy Distillation} (HOPD), which extends teacher distribution matching to every masked position of each denoising step. Because hesitations are not equally informative, we further allocate supervision using hindsight from the completed trajectory, placing more weight on positions whose proposals often disagree with the final token and on blocks whose first-step proposals rarely survive. Since both models already produce distributions at all masked positions, HOPD requires no additional forward passes over TOPD. The only extra cost is evaluating the loss at more positions. With SDAR-1.7B and SDAR-4B students distilled from TraDo-8B-Instruct, HOPD achieves the best average score among the evaluated methods on five math and coding benchmarks, under both static and dynamic decoding and at both scales. It also speeds up decoding: on SDAR-4B, the HOPD student hesitates less and commits 11\% more tokens per denoising step than TOPD, while reaching higher accuracy.
\end{abstract}

\section{Introduction}
Much of the reasoning ability of large language models (LLMs) is acquired in
post-training~\citep{guo2025deepseek,chu2025sft}. Although diffusion large
language models (dLLMs)~\citep{sahoo2024simple,nie2026large,arriola2025block,ye2508dream,cheng2026sdar}
have emerged as a competitive non-autoregressive alternative, post-training for strong reasoning remains challenging, largely because the current
recipes fail to jointly satisfy the desiderata of an effective training signal:
dense supervision at every token and on-policy states drawn from the model's own
generation process. Supervised fine-tuning (SFT) provides dense
supervision by training the model to reconstruct a fixed target response
corrupted with random masks, whereas the randomly masked states it learns from
diverge from the partially decoded states encountered at inference, giving rise
to exposure bias and weaker generalization. Reinforcement learning with
verifiable rewards (RLVR), by contrast, trains on the model's own rollouts and
thereby avoids this mismatch, at the cost of sparse and high-variance
sequence-level rewards that render credit assignment over long denoising
trajectories difficult and optimization expensive~\citep{zhao2025d1,wang2026revolutionizing,ou2026espo}.

On-policy distillation (OPD)~\citep{agarwal2024policy,lu2025onpolicydistillation}
satisfies both desiderata by having the student sample its own outputs and
querying a stronger teacher for a dense token-level signal on exactly those
samples. Unlike an autoregressive student, which makes exactly one prediction
per state, a dLLM student proposes a token for every masked position at each
denoising step and commits only the subset selected by a decoding rule. Therefore,
extending OPD to dLLMs requires deciding at which of these positions the
teacher signal should be applied. Trace-based on-policy distillation
(TOPD)~\citep{ren2026topd} resolves this choice by supervising only the
\emph{trace}, namely the token decisions committed along the trajectory, which
are precisely those that constitute the final response. This supervision scheme is aligned with the underlying decoding process, as the committed tokens are the actions taken by the decoder at each step, and trajectory-level RL for dLLMs likewise evaluates each denoising step only at the tokens it unmasks~\citep{wang2026revolutionizing,huang2025reinforcing,zhan2026simple,wang2026d2,pan2026treerpo}.

However, trace-only supervision leaves out every proposal that the decoder declines
to commit at the current denoising step. We refer to such uncommitted proposals
as \emph{hesitations}, since a proposal whose confidence falls short of the
commitment criterion is withheld much as a speaker hesitates before uttering an
expression~\citep{maclay1959hesitation}. A hesitated position remains masked
and is predicted again at subsequent steps until one of its proposals is
finally committed, so the committed token is only the last in a series of
predictions at that position. TOPD reaches only this last prediction, whereas
supervising hesitations allows the teacher to guide the entire series,
including how quickly it becomes confident enough to be committed. A pilot
study on the rollouts of the base SDAR-4B student confirms that this signal is
substantial, revealing that hesitations account for only 24\% of supervisable
state--position pairs but 66\% of the teacher--student divergence, while the
student already closely matches the teacher at many of the positions
supervised by TOPD. The same study further shows that disagreement varies
systematically with the student's proposal history, being larger on average at
positions whose proposals more often differ from the final token and in blocks
whose first-step proposals less often match the final tokens.

These observations motivate \emph{Hesitation-Aware On-Policy Distillation}
(\ours{}), which extends teacher distribution matching from the committed
positions to every masked position of each visited state. Because both models
already produce distributions at all masked positions, this expansion requires
no forward passes beyond those of TOPD. Guided by the second finding of the
pilot study, \ours{} further allocates supervision within the expanded target
set according to hindsight from the completed trajectory, assigning greater
weight to positions whose proposals more often differ from the final token and
to blocks whose first-step proposals less often match the final tokens. Since
the teacher now shapes the student's predictions at hesitations as well, we
expect \ours{} to raise the student's confidence in teacher-supported tokens,
allowing more positions to reach the commitment threshold earlier and thereby
accelerating decoding. Experiments on SDAR-1.7B and SDAR-4B students
confirm that \ours{} improves both accuracy and decoding efficiency over
TOPD.

We summarize our main contributions as follows:
\begin{itemize}
    \item We identify hesitations as supervisable state--position pairs omitted
    by trace-only distillation, and show in a pilot study that they carry most
    of the teacher--student divergence and that this divergence varies
    systematically with the student's proposal history.
    \item We propose \ours{}, which extends teacher distribution matching to all
masked positions of each visited state without additional forward passes, and
complement it with hindsight weights derived from the completed trajectory so
as to direct more supervision toward positions and blocks where the student's
predictions are least settled.
    \item We evaluate \ours{} against SFT, off-policy distillation,
    trajectory-aware RL, and TOPD on five math and coding benchmarks. \ours{}
    achieves the highest average score under both decoding rules at both
    scales, commits 11\% more tokens per denoising step than TOPD on SDAR-4B, and extends to the full-attention LLaDA-8B-Instruct. 
\end{itemize}
\section{Preliminaries}
\subsection{Masked Diffusion Language Models}
\label{sec:prelim_mdlm}
Let $\mathcal{V}$ be the vocabulary, $\mathrm{[M]} \notin \mathcal{V}$ a
special mask token, and $q$ a prompt. A response state
$s \in (\mathcal{V} \cup \{\mathrm{[M]}\})^{L}$ is a length-$L$ sequence in
which some positions are masked, and we denote the set of these masked
positions by $\mathcal{M}(s)$. A dLLM is a denoiser $\pi_\theta$ that, given
$(q, s)$, outputs in a single forward pass a categorical distribution
$\pi_\theta(\cdot \mid q, s, j)$ over $\mathcal{V}$ for every masked position
$j \in \mathcal{M}(s)$.

\textbf{Block attention.}
Block-attention dLLMs, such as SDAR~\citep{cheng2026sdar} and
TraDo~\citep{wang2026revolutionizing}, partition the response into consecutive
blocks of size $B$ and apply a block-causal attention mask, under which a
position in block $b$ attends to the prompt and to all positions in blocks
$\le b$, but to none in later blocks. Generation therefore proceeds block by
block, with each block starting fully masked, being denoised in at most $K$
steps, and then being frozen before the next block begins. Because attention
within a block is bidirectional, its $B$ positions can be revealed in any
order, which the decoding rule determines at inference time. In the block-wise
objectives below, we slightly abuse $\mathcal{M}(s)$ to denote only the masked
positions of the current block.

\textbf{Decoding rules.}
At denoising step $t$, given the current state $s_t$, the model proposes a
token $\hat{x}_{t,j} \sim \pi_\theta(\cdot \mid q, s_t, j)$ for every masked
position $j$ of the current block, together with a confidence
$c_{t,j} = \pi_\theta(\hat{x}_{t,j} \mid q, s_t, j)$. A decoding rule then
chooses which proposals to \emph{commit}, and the committed tokens are written
into the state. \emph{Static} decoding commits the $B/K$ highest-confidence
positions at each step, assuming $K$ divides $B$
(Appendix~\ref{app:protocol}), whereas \emph{dynamic} decoding commits every
position with $c_{t,j} > \tau$ and falls back to the highest-confidence
position when none qualifies, so that the number of commits varies across
steps. Under either rule, the masked positions of the current block are
partitioned as $\mathcal{M}(s_t) = A_t \cup R_t$ into a \emph{committed} set
$A_t$ and an \emph{uncommitted} set $R_t$, whose positions remain masked and
are proposed again at the next step.

\subsection{On-Policy Distillation}
\label{sec:prelim_opd}
On-policy distillation (OPD)~\citep{agarwal2024policy} is a post-training
method in which a student model learns from its own rollouts rather than from
teacher-generated trajectories. For an autoregressive student $\pi_\theta$ and a
frozen teacher $\pi_{\mathrm{tea}}$, OPD samples a response
$y \sim \pi_\theta(\cdot \mid q)$ and minimizes a token-level divergence between
the next-token distributions of the two models along the same prefixes. Recent
OPD methods~\citep{gu2024minillm,lu2025onpolicydistillation} adopt the reverse
KL divergence, whose mode-seeking behavior encourages the student to
concentrate on the modes preferred by the teacher:
\begin{equation}
  \mathcal{L}_{\mathrm{OPD}}(\theta)
  = \mathbb{E}_{q,\, y \sim \pi_\theta}
    \sum_{i}
    D_{\mathrm{KL}}\!\left(
      \pi_\theta(\cdot \mid q, y_{<i}) \,\|\,
      \pi_{\mathrm{tea}}(\cdot \mid q, y_{<i})
    \right).
  \label{eq:opd}
\end{equation}
When evaluating the divergence over the full vocabulary is costly, the reverse
KL admits a single-sample score-function estimator~\citep{lu2025onpolicydistillation}
which, treating the sampled prefixes as fixed, uses only the sampled token
$x = y_i$:
\begin{equation}
  \nabla_\theta \mathcal{L}
  \approx
  - \nabla_\theta \log \pi_\theta(x \mid q, y_{<i})\;
  \operatorname{sg}\!\big[\log \pi_{\mathrm{tea}}(x \mid q, y_{<i})
                          - \log \pi_\theta(x \mid q, y_{<i})\big],
  \label{eq:k1}
\end{equation}
where $\operatorname{sg}$ denotes the stop-gradient operator.

Given the prefix $y_{<i}$, \Eqref{eq:opd} evaluates the divergence only at the next-token position. In a dLLM, however, each denoising step produces distributions over all masked positions, while the decoder commits only a subset. Extending OPD to dLLMs therefore requires specifying a \emph{target set} of positions for each visited state.

\textbf{Trace-Based OPD for dLLMs.}
For a dLLM rollout, the decoding trajectory is $\xi = (s_0, s_1, \dots, s_T)$: $s_0$ is
fully masked, $s_T$ is the final response, and step $t$ moves from $s_t$
to $s_{t+1}$ by writing the proposals at the committed positions $A_t$ into the state. Trace-Based OPD (TOPD)~\citep{ren2026topd} evaluates the teacher and the student on the same pre-action state $s_t$, i.e., the state before the commitment at step $t$ is applied, and at the same position, and matches them with the reverse KL. Specifically, the loss of TOPD is given by:
\begin{equation}
  \mathcal{L}_{\mathrm{TOPD}}(\theta)
  = \mathbb{E}_{q,\,\xi \sim \pi_\theta}
    \sum_{t:\, |A_t| > 0}
    \frac{1}{|A_t|}
    \sum_{j \in A_t}
    D_{\mathrm{KL}}\!\left(
      \pi_\theta(\cdot \mid q, s_t, j) \,\|\,
      \pi_{\mathrm{tea}}(\cdot \mid q, s_t, j)
    \right).
  \label{eq:topd}
\end{equation}
In this paired objective, the teacher signal is dense at the token level and aligned with the student’s diffusion trace. Only committed proposals are written into the state and thus form the
trajectory, so $A_t$ is a natural target set. TOPD inherits this trace principle from TraceRL~\citep{wang2026revolutionizing}.

\section{Where Does the Teacher Disagree? A Pilot Study}
\label{sec:pilot}

TOPD relies on the decoder's commitment decisions to determine where teacher
supervision is applied. In this section, we first measure the disagreement
that this choice leaves out, and then ask whether simple statistics of the
completed trajectory can identify positions and blocks with greater average
disagreement. The two analyses motivate, respectively, the expanded target set
of \ours{} and its hindsight-based weighting.

\textbf{Hesitations.}
Since both the student and the teacher output a distribution over the
vocabulary at every masked position $j \in \mathcal{M}(s_t) = A_t \cup R_t$,
each such state--position pair is \emph{supervisable} regardless of whether its
proposal is committed, which makes commitment and supervision two distinct
decisions. We call a supervisable pair $(s_t, j)$ with $j \in R_t$ a
\emph{hesitation}, as its proposal is left uncommitted and the student predicts
the same position again at a later step. Letting $\hat{x}_{t,j}$ denote the
sampled proposal and $s_{T,j}$ the final token at position $j$, we partition
the supervisable pairs into three categories:
\begin{itemize}
  \item \textbf{committed}: $j \in A_t$, so the proposal is written into the
    state and becomes the final token. TOPD supervises exactly these pairs.
  \item \textbf{deferred}: $j \in R_t$ and $\hat{x}_{t,j} = s_{T,j}$, so the
    uncommitted proposal matches the token eventually committed there.
  \item \textbf{retracted}: $j \in R_t$ and $\hat{x}_{t,j} \neq s_{T,j}$, so
    the uncommitted proposal differs from the token eventually committed.
\end{itemize}
Deferred and retracted pairs constitute the hesitations. The two kinds of hesitation call for opposite behavior. A deferred proposal already matches the final token, so ideally the student would commit it sooner. A retracted proposal is later abandoned, so remaining uncertain there is appropriate.

\begin{figure}[!t]
  \centering
  \includegraphics[width=\linewidth]{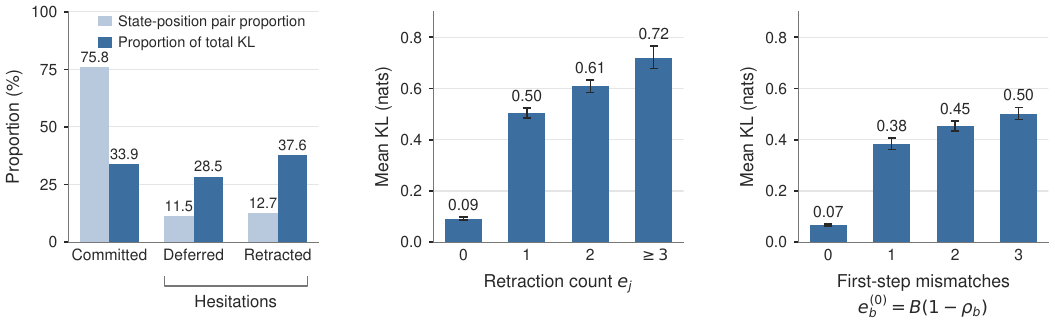}
  \caption{Hesitation pilot study on the rollouts of the base SDAR-4B student
under dynamic decoding, with TraDo-8B-Instruct as the teacher.
\textbf{Left}: share of all supervisable state--position pairs (light blue) and
share of the total teacher--student reverse KL (dark blue) in each commitment
category. \textbf{Middle}: mean per-position KL, averaged over each position's
recorded steps and grouped by the retraction count $e_j$, i.e., the number of
uncommitted proposals at position $j$ that differ from its final token.
\textbf{Right}: mean per-block KL, averaged over masked positions within each
step and then over steps, grouped by the first-step mismatch count
$e_b^{(0)}$, i.e., the number of positions in block $b$ whose first proposal
differs from the final token; only complete four-position blocks are included. Error bars in the middle and right panels are 95\% confidence intervals.}
  \label{fig:pilot}
\end{figure}

\textbf{Setup of our pilot study.}
We sample one trajectory per prompt from base SDAR-4B-Chat on the MATH training prompts, using dynamic decoding with the training settings
(Appendix~\ref{app:details}). At each recorded pre-action state, we
evaluate the student and the frozen TraDo-8B-Instruct teacher on the
same masked positions and measure their KL divergence.  Figure~\ref{fig:pilot} summarizes the results,
and Appendix~\ref{app:pilot} gives the experimental details.

\textbf{Hesitations carry most of the divergence.}
As shown in Figure~\ref{fig:pilot} (left), committed pairs account for
$75.8\%$ of all recorded state--position pairs, whereas deferred and retracted
pairs account for only $11.5\%$ and $12.7\%$, respectively. Despite comprising less than a quarter of the pairs, the two hesitation categories carry $66.1\%$ of the total teacher--student reverse KL. Trace-only supervision therefore omits a disproportionate share of the disagreement,
located precisely at positions that the student must predict again before committing a token.

\textbf{Hindsight and disagreement.}
We next ask whether the student's proposal history, which becomes available
once a trajectory is complete, indicates where the disagreement is
concentrated. At the position level, we define the \emph{retraction count}
$e_j$ as the number of sampled proposals at position $j$ that differ from its
final token, thereby counting disagreements with the eventual outcome rather
than changes between successive proposals. As shown in Figure~\ref{fig:pilot}
(middle), the mean KL of a position, averaged over its recorded steps, grows
from $0.09$ for positions without retractions to $0.50$, $0.61$, and $0.72$
for one, two, and three or more retractions. At the block level, we define the
\emph{first-step retention} $\rho_b$ as the fraction of positions in block $b$
whose first proposal matches the final token, regardless of whether that
proposal is committed, and observe an analogous trend in
Figure~\ref{fig:pilot} (right), where the mean KL of a block rises from $0.07$
to $0.38$, $0.45$, and $0.50$ as the first-step mismatch count
$e_b^{(0)} = B(1-\rho_b)$ increases from zero to three. Since both statistics
average the KL over denoising steps rather than summing it, these trends do not
merely reflect a larger KL sum accumulated over more steps.

Taken together, these observations motivate the two key points:
extending supervision to hesitations, and using hindsight from the completed trajectory to allocate this supervision across positions and blocks. We treat the hindsight statistics as heuristic signals for this allocation and evaluate their effect on training separately in Section~\ref{sec:ablation}.

\section{Hesitation-Aware On-Policy Distillation}
\label{sec:method}

\begin{figure}[!t]
  \centering
  \includegraphics[
    width=0.95\linewidth, trim=50 20 45 60, clip]{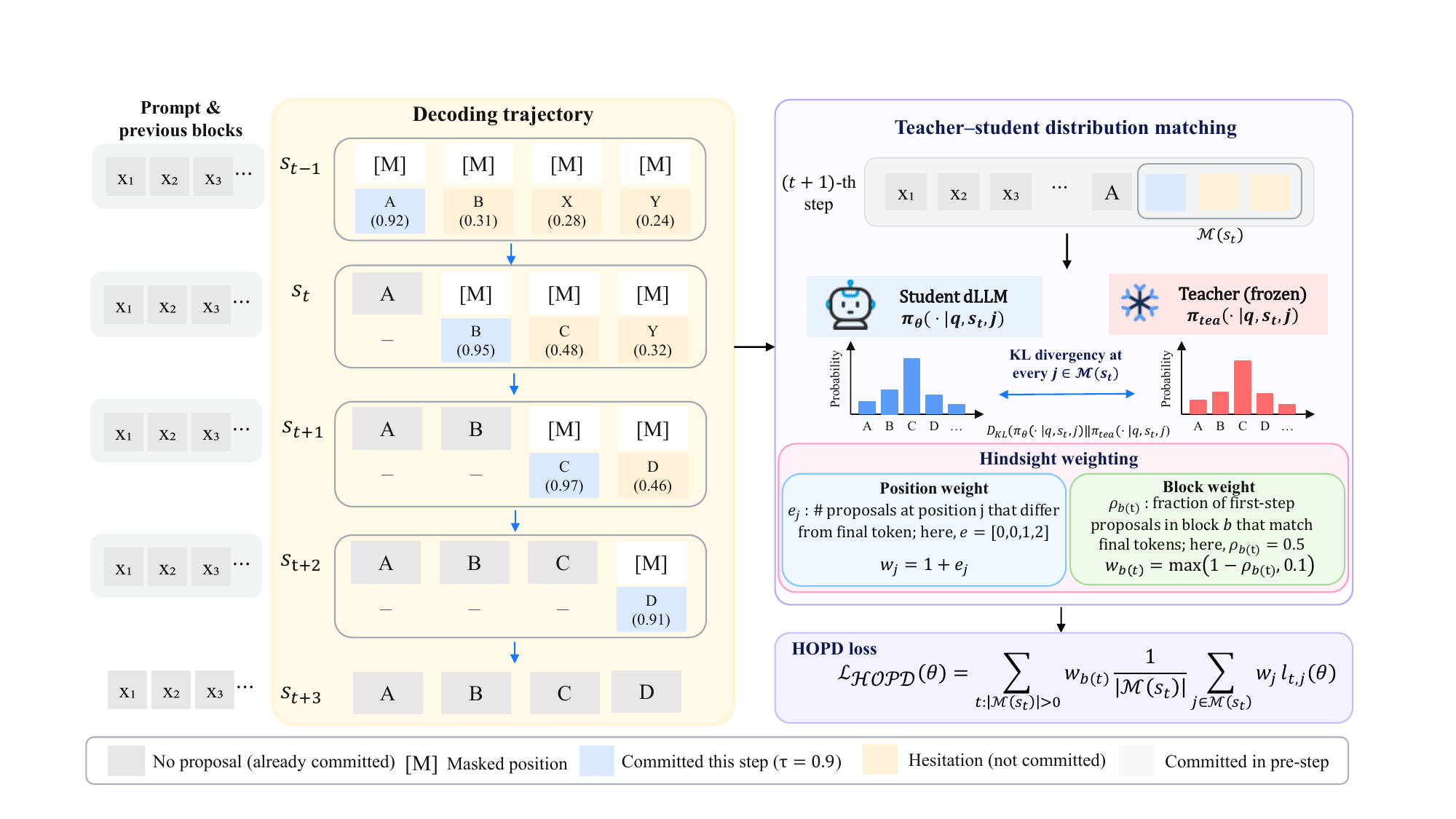}
  \caption{Overview of \ours{}. \textbf{Left}: Dynamic decoding of a
  four-token block ($\tau=0.9$). Blue proposals are committed, while
  yellow proposals are hesitations whose positions remain masked.
  \textbf{Right}: At each visited pre-action state, the student matches
  the frozen teacher's distributions at all masked positions, including
  hesitations. Position and block
  weights are computed from the completed trajectory's retraction counts
  and first-step retention, respectively.}
  \label{fig:framework}
\end{figure}
Building on these findings,  we propose \textit{Hesitation-Aware On-Policy Distillation} (HOPD),  which extends teacher distribution matching from the committed positions to every masked position of each visited state, and weights this supervision using hindsight from the completed trajectory. We first describe how training states are collected, then present the objective
together with its per-position loss and hindsight weights, and finally discuss
its implications for inference.

\textbf{On-policy states.}
At each rollout round, the current student generates trajectories that define the visited pre-action states $s_t$. At each such state, the student is matched to the frozen teacher, while the sampled trajectories remain fixed during the subsequent updates.

\textbf{Hindsight weighting.}
Motivated by the pilot study, we define position and block weights using
the retraction count $e_j$ and first-step retention $\rho_b$ introduced
in Section~\ref{sec:pilot}:
\begin{equation}
  w_j=1+e_j,
  \qquad
  w_b=\max(1-\rho_b,\rho_{\min}),
  \qquad \rho_{\min}=0.1.
  \label{eq:hindsight_weights}
\end{equation}
These weights emphasize positions with more retractions and blocks
with lower first-step retention. Intuitively, $w_j$ reflects hesitation
before settling on a choice: how often the student proposes alternatives
to the token eventually committed at a position. In contrast, $w_b$
reflects the fragility of first impressions: the fraction of a block's
initial proposals that do not survive in the final response.
The floor $\rho_{\min}$ ensures
that every block retains a positive weight.
Both weights are computed from the completed trajectory and shared across the training rows.

\textbf{Training objective.}
For each row, \ours{} matches the student and the frozen teacher on the same state at every masked position, $\mathcal{M}(s_t)=A_t\cup R_t$. The objective is given by:
\begin{equation}
  \mathcal{L}_{\ours}(\theta)
  = \mathbb{E}_{q,\,\xi\sim\pi_\theta}
    \sum_{t:\,|\mathcal{M}(s_t)|>0}
    w_{b(t)}\,
    \frac{1}{|\mathcal{M}(s_t)|}
    \sum_{j\in\mathcal{M}(s_t)}w_j\,\ell_{t,j}(\theta),
  \label{eq:final}
\end{equation}
where $\ell_{t,j}$ is the per-position distribution-matching loss,
$b(t)$ is the block denoised at step $t$. Figure~\ref{fig:framework} illustrates how \ours{} combines
student-generated trajectories, teacher distribution matching,
and hindsight weighting.

\textbf{Full-vocabulary reverse KL.}
For the per-position loss $\ell_{t,j}$, we use full-vocabulary
reverse KL to directly match the student and teacher distributions.
Rather than estimating the divergence from a single sampled token,
we evaluate it over the entire vocabulary at each recorded state
and position:
\begin{equation}
  \ell_{t,j}(\theta)
  = D_{\mathrm{KL}}\!\left(
      \pi_\theta(\cdot \mid q, s_t, j) \,\|\,
      \pi_{\mathrm{tea}}(\cdot \mid q, s_t, j)
    \right),
  \qquad j \in \mathcal{M}(s_t).
  \label{eq:hes}
\end{equation}
This removes token-sampling noise from the per-position divergence
evaluation, while the reverse-KL direction penalizes probability mass
that the student assigns to tokens with little teacher support.
Neither the uncommitted proposal nor the student's
final token is used as a hard target: the target is the teacher's
distribution at that state and position.

\textbf{Inference.}
\ours{} does not change the decoding rule. During the training process,
teacher-supported predictions may reach the confidence threshold
earlier, allowing more commitments per step at the inference stage. Section~\ref{sec:main_results} evaluates the resulting decoding efficiency empirically.

\section{Experiments}
\subsection{Setup}
\label{sec:setup}

\textbf{Models and data.}
Main experiments use SDAR-4B-Chat and SDAR-1.7B-Chat~\citep{cheng2026sdar} as
students and TraDo-8B-Instruct~\citep{wang2026revolutionizing} as the teacher. For
mathematical post-training, we use the MATH training set~\citep{hendrycks2021measuring},
retaining level 3--5 problems, yielding 8K
tasks. For the coding setting, we use 6K verified problems from PrimeIntellect~\citep{jaghouar2024intellect}, verified by
DeepCoder~\citep{deepcoder2025}.

\textbf{Rollout and training.}
Training rollouts use dynamic decoding with threshold $\tau = 0.9$ and
temperature $1.0$. A \emph{rollout round} samples one response for each
of 64 prompts with the JetEngine inference engine~\citep{cheng2026sdar},
synchronized with the live student before every round, and applies one
epoch of training on the resulting trace rows. A round thus contains
multiple optimizer steps rather than a single update.
Appendix~\ref{app:training} lists every hyperparameter.

\textbf{Baselines.}
\emph{SFT} is semi-autoregressive supervised fine-tuning~\citep{wang2026revolutionizing}
on fixed teacher responses: every 4-token block of a TraDo-8B-Instruct
response is corrupted with an independent mask rate drawn from
$[0.1, 0.9]$ and reconstructed with a block-wise cross entropy. \emph{Off-policy} applies the full-vocabulary reverse KL at the
committed positions of frozen teacher trajectories instead of the current
student's~\citep{ren2026topd}.
\emph{TraceRL}~\citep{wang2026revolutionizing} is trajectory-aware RL with verifiable
rewards. For SDAR-4B, we evaluate the released TraDo-4B-Instruct, and for
SDAR-1.7B we train with the public implementation.
\emph{TOPD}~\citep{ren2026topd} uses the sampled-token reverse-KL estimator. We also evaluate a full-vocabulary reverse-KL variant of TOPD in Appendix~\ref{app:topd_full_rkl}. Appendix~\ref{app:baselines} gives the baseline configurations.

\textbf{Evaluation.}
We report MATH500~\citep{lightman2023verify}, AIME2024~\citep{maa2024aime},
and GSM8K~\citep{cobbe2021gsm8k} for mathematics, and
LiveCodeBench-v2~\citep{jain2024livecodebench} and
LiveBench~\citep{white2024livebench} for coding, under both static and dynamic ($\tau = 0.9$)
following the evaluation setting with
TraceRL~\citep{wang2026revolutionizing}. MATH500, GSM8K, and the coding benchmarks
report avg@3; AIME2024 reports avg@20. Checkpoints are selected on MATH500 or LiveCodeBench-v2 under each decoding rule (Appendix~\ref{app:selection}).

\subsection{Main Results}
\label{sec:main_results}

\begin{table*}[!t]
    \centering
    \caption{The main benchmark results across different math and coding tasks. \enquote{Static} refers to static sampling, and \enquote{Dynamic} refers to dynamic sampling.}
    \label{tab:main_results}
    \small
    \setlength{\tabcolsep}{4pt}
    \resizebox{\linewidth}{!}{%
    \begin{tabular}{l cc cc cc cc cc cc}
        \toprule
        \multirow{2}{*}{\textbf{Model}}
        & \multicolumn{2}{c}{\textbf{MATH500}}
        & \multicolumn{2}{c}{\textbf{AIME2024}}
        & \multicolumn{2}{c}{\textbf{GSM8K}}
        & \multicolumn{2}{c}{\textbf{LiveCodeBench-v2}}
        & \multicolumn{2}{c}{\textbf{LiveBench}}
        & \multicolumn{2}{c}{\textbf{Avg.}}
        \\
        \cmidrule(lr){2-3}
        \cmidrule(lr){4-5}
        \cmidrule(lr){6-7}
        \cmidrule(lr){8-9}
        \cmidrule(lr){10-11}
        \cmidrule(lr){12-13}
        & \textbf{Static} & \textbf{Dynamic}
        & \textbf{Static} & \textbf{Dynamic}
        & \textbf{Static} & \textbf{Dynamic}
        & \textbf{Static} & \textbf{Dynamic}
        & \textbf{Static} & \textbf{Dynamic}
        & \textbf{Static} & \textbf{Dynamic}
        \\
        \midrule

        TraDo-8B-Instruct (teacher)
        & 78.5 & 75.5
        & 13.3 & 11.0
        & 92.3 & 91.2
        &25.9  & 22.4
        & 22.7 & 20.6
        & 46.5 & 44.1 \\
        \midrule

        SDAR-1.7B-Chat
        & 61.7 & 54.2
        & 4.0 & 4.7
        & 80.9 & 78.4
        & 7.2 & 4.2
        & 5.2 & 4.2
        & 31.8 & 29.1 \\

        \quad + SFT
        & 49.2 & 46.3
        & 1.0 & 3.8
        & 77.0 & 74.7
        & 7.8 & 6.6
        & 6.2 & 8.6
        & 28.2 & 28.0 \\

        \quad + Off-policy
        & 63.3 & 56.6
        & \textbf{6.8} & 3.2
        & 81.6 & 75.1
        & \textbf{11.3} & \textbf{9.1}
        & \textbf{11.7} & 7.8
        & 34.9 & 30.4 \\

        \quad + TraceRL
        & 62.5 & 56.1
        & 6.7 & 3.5
        & 81.1 & 78.7
        & 9.4 & 4.7
        & 10.7 & 5.7
        & 34.1 & 29.7 \\

        \quad + TOPD
        & 64.5 & 58.1
        & 2.2 & 3.2
        & 81.9 & 78.2
        & 10.4 & 7.7
        & 9.4 & 6.2
        & 33.7 & 30.7 \\

        \rowcolor{gray!15}
        \quad + \ours{} (ours)
        & \textbf{66.8} & \textbf{59.5}
        & 3.7 & \textbf{5.0}
        & \textbf{83.6} & \textbf{78.8}
        & 11.1 & 8.9
        & 10.9 & \textbf{9.9}
        & \textbf{35.2} & \textbf{32.4} \\

        \midrule
        SDAR-4B-Chat
        & 70.5 & 66.1
        & 7.5 & 7.8
        & 90.5 & 88.9
        & 19.3 & 11.7
        & 19.0 & 9.6
        & 41.4 & 36.8 \\

        \quad + SFT
        & 60.7 & 53.3
        & 5.7 & 4.8
        & 83.7 & 81.7
        & 12.9 & 11.0
        & 12.8 & 10.9
        & 35.1 & 32.3 \\

        \quad + Off-policy
        & 75.6 & 71.3
        & 7.8 & 8.0
        & 91.3 & 90.1
        & 21.5 & 19.3
        & 19.3 & 17.2
        & 43.1 & 41.2 \\

        \quad + TraceRL (TraDo-4B-Instruct)
        & \textbf{76.2} & 71.0
        & 7.7 & 8.8
        & 91.3 & 89.7
        & 18.0 & 13.9
        & 14.8 & 9.9
        & 41.6 & 38.7 \\

        \quad + TOPD
        & 74.9 & 71.3
        & 11.2 & 7.2
        & \textbf{92.0} & 90.0
        & \textbf{21.9} & 18.6
        & 19.8 & 17.4
        & 43.9 & 40.9 \\

        \rowcolor{gray!15}
        \quad + \ours{} (ours)
        & 74.9 & \textbf{72.5}
        & \textbf{15.0} & \textbf{9.7}
        & 91.1 & \textbf{90.7}
        & 20.9 & \textbf{19.8}
        & \textbf{22.1} & \textbf{19.8}
        & \textbf{44.8} & \textbf{42.5} \\

        \bottomrule
    \end{tabular}}
\end{table*}

\textbf{Main results.}
Table~\ref{tab:main_results} shows that \ours{} achieves the highest
five-benchmark average among the evaluated training methods at both model scales under both decoding rules. Relative to TOPD, the average
gains are $1.5$/$1.7$ percentage points on SDAR-1.7B and $0.9$/$1.6$ on SDAR-4B under static/dynamic decoding, respectively.
Moreover, the advantage of \ours{} is particularly consistent under dynamic decoding: On SDAR-1.7B, it also outperforms TOPD on every benchmark under both decoding rules. On SDAR-4B, its static AIME2024 and
LiveBench scores are the highest, improving over TOPD by $3.8$ and
$2.3$ points, respectively. These results demonstrate the effectiveness of \ours{} across model scales and decoding strategies.

\begin{figure}[t]
  \centering
  \includegraphics[width=0.9\linewidth]{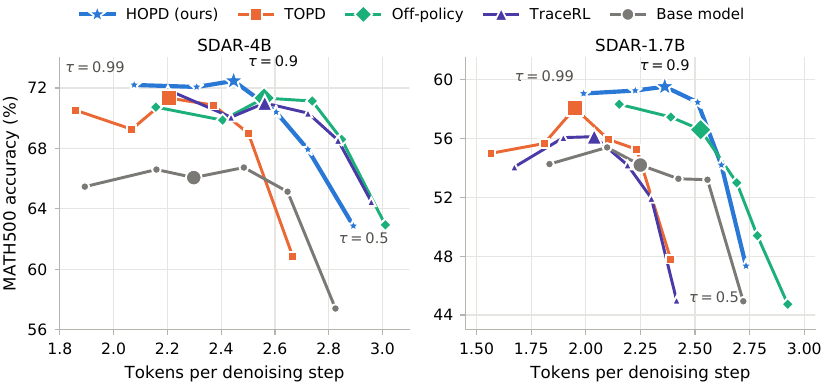}
  \caption{Accuracy against committed tokens per denoising step on
  MATH500 (dynamic decoding) as the commit threshold $\tau$ goes from
  $0.5$ to $0.99$, right to left along each curve; large markers denote
  the training threshold $\tau = 0.9$. \textbf{Left}: SDAR-4B. \textbf{Right}: SDAR-1.7B.}
  \label{fig:speed}
\end{figure}

\textbf{Decoding speed.} To evaluate the decoding efficiency of students trained with \ours{}, we sweep the commit threshold $\tau \in \{0.5, 0.7, 0.8, 0.9, 0.95, 0.99\}$ during dynamic decoding and measure the average number of committed tokens per denoising step.
Figure~\ref{fig:speed} shows that \ours{} achieves a better
accuracy--efficiency trade-off than TOPD on MATH500 at both model
scales. On SDAR-4B, this advantage is particularly clear in the high-accuracy regime: at every evaluated threshold $\tau \geq 0.9$, the \ours{}-trained student achieves the highest accuracy among the compared methods.
At the training threshold $\tau = 0.9$, it reaches $72.5\%$ accuracy versus $71.3\%$ for TOPD, while committing $11\%$ more tokens per denoising step.
Wall-clock throughput results are reported in Appendix~\ref{app:wallclock}.

\Needspace{16\baselineskip}
\begin{wraptable}{r}{0.48\textwidth}
  \centering
  \setlength{\abovecaptionskip}{0pt}
  \setlength{\belowcaptionskip}{6pt}
  \caption{Commitment categories after training.}
  \label{tab:pilot_after}
  \small
  \setlength{\tabcolsep}{4pt}
  \begin{tabular}{@{}lccc@{}}
    \toprule
    & \textbf{Base} & \textbf{TOPD} & \textbf{\ours{}} \\
    \midrule
    \multicolumn{4}{@{}l}{\textit{Proportion of pairs}} \\
    Committed & 75.2\% & 72.6\% & 77.1\% \\
    Deferred  & 11.9\% & 12.0\% & 10.5\% \\
    Retracted & 12.9\% & 15.4\% & 12.4\% \\
    \midrule
    \multicolumn{4}{@{}l}{\textit{Predictive entropy (nats)}} \\
    Committed & 0.141 & 0.177 & 0.124 \\
    Deferred  & 0.962 & 1.173 & 1.017 \\
    Retracted & 1.433 & 1.984 & 1.615 \\
    \bottomrule
  \end{tabular}
\end{wraptable}
\textbf{Revisiting hesitations after training.}
We revisit the pilot analysis using the math-trained SDAR-4B students. 
Table~\ref{tab:pilot_after} shows that TOPD shifts pairs from committed to retracted and raises the entropy in every category, so the student becomes less certain overall.
In contrast, \ours{} has the largest committed proportion and the smallest
deferred and retracted shares, and its entropy is lower than TOPD's in every category. At deferred pairs in particular, \ours{} largely avoids the loss of certainty under TOPD ($1.017$ vs.\ $1.173$ nats, against $0.962$ for the base student), and such pairs are rarer than even for the base student. Fewer deferred pairs mean fewer proposals that already match the final token but are held back, so more positions are committed at each step, consistent with the faster decoding in Figure~\ref{fig:speed}.

\begin{table}[!t]
    \centering
    \caption{Full-attention model: LLaDA-8B-Instruct with ESPO-trained
    teachers.}
    \label{tab:full_attention}
    \small
    \setlength{\tabcolsep}{4pt}
    \resizebox{\linewidth}{!}{%
    \begin{tabular}{l cc cc cc cc cc cc}
        \toprule
        \multirow{2}{*}{\textbf{Model}}
        & \multicolumn{2}{c}{\textbf{MATH500}}
        & \multicolumn{2}{c}{\textbf{AIME2024}}
        & \multicolumn{2}{c}{\textbf{GSM8K}}
        & \multicolumn{2}{c}{\textbf{LiveCodeBench-v2}}
        & \multicolumn{2}{c}{\textbf{LiveBench}}
        & \multicolumn{2}{c}{\textbf{Avg.}}
        \\
        \cmidrule(lr){2-3}
        \cmidrule(lr){4-5}
        \cmidrule(lr){6-7}
        \cmidrule(lr){8-9}
        \cmidrule(lr){10-11}
        \cmidrule(lr){12-13}
        & \textbf{Static} & \textbf{Dynamic}
        & \textbf{Static} & \textbf{Dynamic}
        & \textbf{Static} & \textbf{Dynamic}
        & \textbf{Static} & \textbf{Dynamic}
        & \textbf{Static} & \textbf{Dynamic}
        & \textbf{Static} & \textbf{Dynamic}
        \\
        \midrule
        LLaDA-8B-Instruct
        & 36.8 & 37.4
        & 0.0 & 0.0
        & 81.8 & 82.2
        & 6.5 & 6.6
        & 7.3 & 7.8
        & 26.5 & 26.8 \\

        \quad + ESPO (teacher)
        & 37.4 & 36.0
        & 0.0 & 0.0
        & 83.3 & 82.8
        & 8.4 & 7.3
        & 12.8 & 12.0
        & 28.4 & 27.6 \\
        \midrule
        \quad + TOPD
        & 38.2 & \textbf{38.3}
        & \textbf{0.3} & \textbf{0.3}
        & \textbf{83.8} & 82.2
        & 6.8 & \textbf{6.1}
        & 10.2 & 9.4
        & 27.9 & 27.3 \\

        \rowcolor{gray!15}
        \quad + \ours{} (ours)
        & \textbf{39.0} & 38.1
        & 0.2 & 0.2
        & 82.2 & \textbf{82.4}
        & \textbf{7.7} & \textbf{6.1}
        & \textbf{13.0} & \textbf{10.7}
        & \textbf{28.4} & \textbf{27.5} \\
        \bottomrule
    \end{tabular}}
\end{table}

\section{Discussion and Ablation Studies}
\label{sec:ablation}
\textbf{Generalization to full-attention models.}
To assess the applicability of \ours{} to full-attention dLLMs,
we apply it to LLaDA-8B-Instruct, using its ESPO-trained math
and code variants~\citep{ou2026espo} as the respective teachers.
Training uses 32-token blocks with a 128-token future horizon.
Detailed experimental settings are provided in
Appendix~\ref{app:llada}.
As shown in Table~\ref{tab:full_attention}, \ours{} achieves
higher average scores than TOPD under static
($28.4$ vs.\ $27.9$) and dynamic ($27.5$ vs.\ $27.3$) decoding.
For example, on LiveBench under static decoding, \ours{} scores
$13.0$ compared with $10.2$ for TOPD.
These results support the applicability of \ours{} to the full-attention LLaDA model.

\begin{table}[t]
  \centering
  \caption{Comparison of divergence objectives on math with SDAR-4B-Chat.}
  \label{tab:divergence}
  \small
  \setlength{\tabcolsep}{4pt}
  \resizebox{\linewidth}{!}{%
  \begin{tabular}{l cccc cccc}
    \toprule
    & \multicolumn{4}{c}{\textbf{Static}}
    & \multicolumn{4}{c}{\textbf{Dynamic}} \\
    \cmidrule(lr){2-5}\cmidrule(lr){6-9}
    \textbf{Per-position loss}
    & MATH500 & AIME2024 & GSM8K & Avg.
    & MATH500 & AIME2024 & GSM8K & Avg. \\
    \midrule
    Full forward KL
    & 75.6 & 9.8 & 91.1 & 58.8
    & 71.1 & 8.8 & 90.3 & 56.7 \\

    Full JSD ($\beta=0.5$)
    & \textbf{75.9} & 8.2 & 91.5 & 58.5
    & 71.8 & 8.5 & 90.6 & 57.0 \\

    Sampled-token reverse KL
    & 75.3 & 7.8 & \textbf{91.7} & 58.3
    & 71.9 & 8.3 & \textbf{91.0} & 57.1 \\

    \rowcolor{gray!15}
    Full reverse KL
    & 74.9 & \textbf{15.0} & 91.1 & \textbf{60.3}
    & \textbf{72.5} & \textbf{9.7} & 90.7 & \textbf{57.6} \\
    \bottomrule
  \end{tabular}}
\end{table}

\textbf{Effect of divergence objectives.}
A key design choice in \ours{} is the divergence used for per-position distribution matching between the frozen teacher and the student. We compare three full-vocabulary objectives (forward KL, reverse KL, and Jensen--Shannon divergence, JSD) and sampled-token reverse KL, on math tasks with SDAR-4B-Chat in Table~\ref{tab:divergence}. Experimental settings are provided in
Appendix~\ref{app:divergence}.  Full
reverse KL has the highest reported three-benchmark average in this
comparison ($60.3$ static, $57.6$ dynamic). In contrast, other objectives provide limited improvements. We therefore adopt full-vocabulary reverse KL as the default objective.

\begin{table}[t]
  \centering
  \caption{Comparison of hindsight weights on math. Bold: best per column.}
  \label{tab:weights}
  \small
  \setlength{\tabcolsep}{4pt}
  \resizebox{\linewidth}{!}{%
  \begin{tabular}{l cccc cccc}
    \toprule
    & \multicolumn{4}{c}{\textbf{Static}}
    & \multicolumn{4}{c}{\textbf{Dynamic}} \\
    \cmidrule(lr){2-5}\cmidrule(lr){6-9}
    \textbf{Weights}
    & MATH500 & AIME2024 & GSM8K & Avg.
    & MATH500 & AIME2024 & GSM8K & Avg. \\
    \midrule
    Uniform ($w_j = w_b = 1$)
    & \textbf{75.5} & 11.2 & 91.2 & 59.3
    & \textbf{72.5} & 9.7 & 90.2 & 57.5 \\
    $w_j$ only ($w_b=1$)
    & 75.3 & 8.5 & 91.5 & 58.4
    & 71.5 & 8.5 & \textbf{90.8} & 56.9 \\
    $w_b$ only ($w_j=1$)
    & \textbf{75.5} & 7.8 & \textbf{92.0} & 58.4
    & 72.3 & \textbf{10.0} & 90.5 & \textbf{57.6} \\
    \rowcolor{gray!15}
    $w_j$ and $w_b$ (\ours{})
    & 74.9 & \textbf{15.0} & 91.1 & \textbf{60.3}
    & \textbf{72.5} & 9.7 & 90.7 & \textbf{57.6} \\
    \bottomrule
  \end{tabular}}
\end{table}

\textbf{Discussion on hindsight weights.}
We evaluate whether weighting by trajectory hesitation provides further gains over uniform supervision of all masked positions. Table~\ref{tab:weights} compares four weighting configurations with the all-masked target set and full-vocabulary reverse KL fixed. 
On SDAR-4B, combining position and block weights improves the static average from $59.3$ to $60.3$, with AIME2024 increasing from $11.2$ to $15.0$, while the dynamic average improves slightly from $57.5$ to $57.6$.
Similar average gains over uniform weighting hold for SDAR-1.7B math and for code at both scales (Appendix~\ref{app:weights}).
These results support using trajectory hesitation to guide the allocation of supervision, beyond simply including all masked positions.

\section{Related Work}
\textbf{Diffusion language models.}
Most current dLLMs are masked diffusion models with an absorbing-state
corruption process~\citep{austin2021d3pm,sahoo2024simple,shi2024simplified,ou2025radd},
trained from scratch~\citep{nie2026large,zhu2025llada15,bie2025llada2} or
from autoregressive weights~\citep{gong2025diffullama,ye2508dream}. Block
diffusion decodes fixed-size blocks left to right with bidirectional
attention within a block~\citep{arriola2025block}. SDAR adapts
autoregressive checkpoints into this form~\citep{cheng2026sdar}, and TraDo
trains them further with RL~\citep{wang2026revolutionizing}. Inference
uses confidence-based parallel decoding with KV
caches~\citep{wu2025fastdllm,ma2025dkvcache}, and several methods revoke,
defer, or correct low-confidence tokens at inference
time~\citep{hong2025wino,shu2026dcd,schiff2026learn,frkovic2026remasking}.
Among post-training methods, SFT reconstructs randomly masked targets and
is sensitive to the mask pattern~\citep{wang2026revolutionizing,ren2026topd};
RL methods must approximate intractable sequence
likelihoods~\citep{zhao2025d1,zhu2025llada15,zhao2025diffpo,wang2026spg,ou2026espo,zhong2026stabilizing},
with MDPO training on decoding states~\citep{he2025mdpo} and TraceRL
assigning credit along the decoding trace~\citep{wang2026revolutionizing},
both from a scalar reward. \ours{} also trains along the decoding trace
but uses the teacher's distribution at every visited state as the signal
and leaves the decoder unchanged.

\textbf{On-policy distillation.}
Knowledge distillation matches a student to a teacher's output
distribution~\citep{hinton2015distilling}. Sequence-level distillation
trains on teacher-generated outputs~\citep{kim2016sequence} and, like SFT,
is subject to exposure bias~\citep{ross2011dagger}. On-policy distillation
instead queries the teacher on student
samples~\citep{agarwal2024policy,gu2024minillm,xu2025skd}. For masked diffusion, TOPD~\citep{ren2026topd}
distills a stronger dLLM teacher on the committed decisions of the
student's decoding trace, and dOPSD~\citep{dat2026dopsd} performs
self-distillation with the student's own trajectory as privileged
information. Unlike TOPD, \ours{} also matches teacher and student distributions at masked positions whose proposals are not committed at the current step, bringing the student's hesitations into on-policy distillation.

\section{Conclusion}
We presented \ours{}, a hesitation-aware on-policy
distillation for diffusion large language models. Our pilot study shows
that positions whose proposals are not committed account for a
disproportionate share of teacher--student disagreement, yet are
omitted by previous methods. Motivated by this finding, \ours{} extends teacher supervision to all masked positions along student rollouts and
uses trajectory hindsight to reweight this supervision, without
changing the decoder or requiring additional model forward passes
for a given rollout. Across five math and coding benchmarks,
\ours{} achieves the highest average scores among the evaluated methods on SDAR-1.7B and SDAR-4B under both static and
dynamic decoding. It also improves the accuracy--efficiency
trade-off on MATH500 and extends to full-attention LLaDA.
These results highlight the value of supervising the student's
hesitations rather than restricting distillation to committed
decisions.

\clearpage

\subsection*{AI use statement}

In this work, we used generative AI tools for no tasks with required disclosure. We have not used generative AI tools for developing conceptual frameworks, proposing hypotheses, designing experiments, implementing methods, translation, generating synthetic data, cleaning datasets, or interpreting results, and formulating mathematical claims, writing proofs, and qualitative data analysis are not applicable to this work. Additionally, we used generative AI tools for editing the manuscript to improve readability and assisting with the design of scientific figures. We have reviewed all AI-assisted work. All AI-suggested text edits were checked by the authors to preserve the original meaning, and all figure content was verified by the authors against our experimental results. We take responsibility for the final content of this work, including text, claims or artifacts produced with the aid of generative AI.

\section*{Reproducibility Statement}
To facilitate reproducibility, we provide detailed implementation settings, hyperparameters, and additional experimental details in the appendix.

\bibliography{iclr2027_conference}
\bibliographystyle{iclr2027_conference}

\clearpage
\appendix
\section{Pilot Study Details}
\label{app:pilot}
\paragraph{Setup.} We use base SDAR-4B-Chat as the student and prompts from the MATH training set. The model generates one trajectory per prompt using the protocol in Section~3, with seed 1234. A category's pair proportion is its number of pairs divided by the total number of recorded supervisable pairs, and its KL proportion is the sum of KL over its pairs divided by the sum over all recorded supervisable pairs.

\textbf{Position- and block-level aggregation.}
Figure~\ref{fig:pilot} uses the base model's trajectories.
Let $\mathcal{T}_j$ be the steps with a KL measurement at position $j$,
and $\mathcal{T}_b$ the recorded denoising
steps of block $b$. The middle and right panels respectively aggregate
\begin{equation*}
  \bar{\ell}_j = \frac{1}{|\mathcal{T}_j|}
    \sum_{t\in\mathcal{T}_j}\ell_{t,j},
  \qquad
  \bar{\ell}_b = \frac{1}{|\mathcal{T}_b|}
    \sum_{t\in\mathcal{T}_b}
    \frac{1}{|\mathcal{M}(s_t)|}
    \sum_{j\in\mathcal{M}(s_t)}\ell_{t,j}.
\end{equation*}

Table~\ref{tab:pilot_allocation} gives the group sizes and means.
The error bars in the middle and right panels are percentile $95\%$ confidence
intervals from $5{,}000$ bootstrap draws with seed $1234$. Each draw
resamples the $256$ prompts with replacement, retaining each sampled
prompt's entire trajectory, and recomputes the group means. Thus,
positions and blocks within a trajectory are not treated as independent
bootstrap units.

\begin{table}[h]
  \centering
  \caption{Statistics behind the middle and right panels of
  Figure~\ref{fig:pilot}. Each unit is a position in the middle panel
  and a complete four-position block in the right panel.}
  \label{tab:pilot_allocation}
  \small
  \setlength{\tabcolsep}{6pt}
  \begin{tabular}{l c r r c c}
    \toprule
    \textbf{Unit} & \textbf{Group} & \textbf{Units} & \textbf{Prompts}
    & \textbf{Mean KL} & \textbf{95\% CI} \\
    \midrule
    Position & $e_j=0$ & 141,254 & 256 & 0.0924 & $[0.0872,0.0979]$ \\
                & $e_j=1$ & 14,588  & 256 & 0.5044 & $[0.4857,0.5245]$ \\
                & $e_j=2$ & 4,545   & 255 & 0.6084 & $[0.5845,0.6341]$ \\
                & $e_j\geq3$ & 1,114 & 223 & 0.7196 & $[0.6786,0.7655]$ \\
    \midrule
    Block & $e_b^{(0)}=0$ & 29,416 & 256 & 0.0672 & $[0.0629,0.0718]$ \\
             & $e_b^{(0)}=1$ & 5,718  & 256 & 0.3824 & $[0.3612,0.4068]$ \\
             & $e_b^{(0)}=2$ & 3,339  & 255 & 0.4530 & $[0.4334,0.4740]$ \\
             & $e_b^{(0)}=3$ & 1,709  & 244 & 0.5021 & $[0.4787,0.5265]$ \\
    \bottomrule
  \end{tabular}
\end{table}

\paragraph{Hesitation and correctness.} For each trajectory, the hesitation rate is the fraction of supervisable pairs whose proposals are not committed. Its mean is higher for incorrect than for correct responses (27.6\% vs.\ 22.8\%), with a bootstrap 95\% confidence interval of [2.9, 6.7] percentage points for the difference. The gap also persists within response-length bins; for example, the rates are 32.5\% vs.\ 23.6\% for responses of 300--600 tokens.

\section{Experimental Details}
\label{app:details}
This appendix shows the details needed to reproduce the experiments of
Section~\ref{sec:setup}: the training configuration shared by the
on-policy distillation runs (Appendix~\ref{app:training}), the
configuration of every method in Table~\ref{tab:main_results}
(Appendix~\ref{app:baselines}), the prompt templates
(Appendix~\ref{app:prompts}), and the evaluation protocol, including
decoding, grading, and checkpoint selection
(Appendix~\ref{app:protocol}).

\subsection{Training Configuration}
\label{app:training}
Table~\ref{tab:hparams} lists the settings shared by the on-policy
distillation runs behind the main results. Rollouts use
dynamic decoding: at each denoising step the engine commits every masked
position of the current block whose sampled-token probability exceeds
$\tau = 0.9$, and the $B/K = 1$ most confident position when none does.
The step budget $K = 4$ equals the block size $B = 4$, so a block closes
in one to four steps and the budget never binds. TOPD and \ours{}
are trained with the same code base, models, data, rollout protocol, and
optimizer; they differ only in the objective, which
Appendix~\ref{app:baselines} specifies method by method together with the
configuration of every other baseline.

\begin{table}[h]
  \centering
  \caption{Training configuration shared by the TOPD and \ours{} runs of
  Table~\ref{tab:main_results}}
  \label{tab:hparams}
  \small
  \setlength{\tabcolsep}{5pt}
  \resizebox{\linewidth}{!}{%
  \begin{tabular}{l l l}
    \toprule
    \textbf{Setting} & \multicolumn{2}{l}{\textbf{Value}} \\
    \midrule
    \multicolumn{3}{l}{\textit{Models}} \\
    Student & \multicolumn{2}{l}{SDAR-4B-Chat or SDAR-1.7B-Chat~\citep{cheng2026sdar}} \\
    Teacher & \multicolumn{2}{l}{TraDo-8B-Instruct~\citep{wang2026revolutionizing}, frozen} \\
    Precision & \multicolumn{2}{l}{bfloat16 weights, bf16 mixed precision} \\
    Gradient checkpointing & \multicolumn{2}{l}{enabled} \\
    Parallelism & \multicolumn{2}{l}{DeepSpeed ZeRO-2, no offload; 4 GPUs (2 GPUs with accumulation for the 1.7B code run)} \\
    \midrule
    \multicolumn{3}{l}{\textit{Data}} \\
    Math prompts & \multicolumn{2}{l}{MATH training set, levels 3--5 (8,523 problems)} \\
    Code prompts & \multicolumn{2}{l}{PrimeIntellect, verified stdio problems (5,954)} \\
    Max prompt length & \multicolumn{2}{l}{784 tokens} \\
    \midrule
    \multicolumn{3}{l}{\textit{Rollout (on-policy sampling)}} \\
    Inference engine & \multicolumn{2}{l}{JetEngine, weights synchronized with the student before every round} \\
    Prompts per round / responses per prompt & \multicolumn{2}{l}{64 / 1} \\
    Block size / denoising steps per block & \multicolumn{2}{l}{4 / 4} \\
    Decoding rule & \multicolumn{2}{l}{dynamic: commit every position with $c_j > 0.9$, top-1 fallback} \\
    Confidence $c_j$ & \multicolumn{2}{l}{probability of the sampled token} \\
    Temperature / top-$p$ / top-$k$ & \multicolumn{2}{l}{1.0 / 1.0 / off} \\
    Max response length & \multicolumn{2}{l}{2000 tokens} \\
    Rollout seed & \multicolumn{2}{l}{1234} \\
    \midrule
    \multicolumn{3}{l}{\textit{Optimization}} \\
    Optimizer & \multicolumn{2}{l}{AdamW, $\beta_1 = 0.9$, $\beta_2 = 0.999$, $\epsilon = 10^{-8}$} \\
    Learning rate & \multicolumn{2}{l}{$2 \times 10^{-7}$, constant} \\
    Weight decay & \multicolumn{2}{l}{0} \\
    Effective batch size & \multicolumn{2}{l}{16 trace rows (4 per GPU $\times$ 4 GPUs $\times$ 1 accumulation)} \\
    Gradient clipping (max norm) & \multicolumn{2}{l}{1.0} \\
    Rollout rounds / epochs per round & \multicolumn{2}{l}{30 / 1} \\
    Checkpoint cadence & \multicolumn{2}{l}{every 5 rounds} \\
    \bottomrule
  \end{tabular}}
\end{table}

\subsection{Baseline Configurations}
\label{app:baselines}
Every baseline method in Table~\ref{tab:main_results} starts from the same student checkpoint, uses the same prompts, prompt template, and response limit, is evaluated with the same pipeline, and, where a teacher is involved, uses the same frozen TraDo-8B-Instruct. Table~\ref{tab:baselines} summarizes
what each method trains on. The paragraphs below give the remaining
settings. Anything not mentioned follows Table~\ref{tab:hparams}.

\begin{table}[h]
  \centering
  \caption{What each baseline method of Table~\ref{tab:main_results} trains on.
  States: inputs on which the loss is evaluated. Positions: response
  positions that receive gradient at each state.}
  \label{tab:baselines}
  \small
  \setlength{\tabcolsep}{4pt}
  \begin{tabular}{@{}l >{\raggedright\arraybackslash}p{2.8cm} >{\raggedright\arraybackslash}p{1.8cm} >{\raggedright\arraybackslash}p{3.0cm} >{\raggedright\arraybackslash}p{3.3cm}@{}}
    \toprule
    \textbf{Method} & \textbf{Training states} & \textbf{Signal} & \textbf{Supervised positions} & \textbf{Per-position loss} \\
    \midrule
    SFT & fixed teacher responses under random block masks & teacher tokens & masked positions of each block & cross entropy \\
    Off-policy & frozen teacher trajectories & teacher distribution & committed positions of the teacher's trace & full-vocabulary reverse KL \\
    TraceRL & student rollouts & verifiable reward & committed positions along the trace & clipped policy gradient with KL penalty \\
    TOPD & student rollouts & teacher distribution & committed positions $A_t$ & sampled-token reverse-KL estimator \\
    \ours{} & student rollouts & teacher distribution & all masked positions $\mathcal{M}(s_t)$ & full-vocabulary reverse KL with hindsight weights \\
    \bottomrule
  \end{tabular}
\end{table}

\textbf{Semi-autoregressive SFT.}
The SFT data are fixed teacher responses, one per training prompt and 
generated with the rollout protocol of Table~\ref{tab:hparams}. Training follows the semi-autoregressive objective of \citet{wang2026revolutionizing}: each 4-token block of the response is corrupted with a random mask rate, and the model is trained with cross entropy on the masked positions under block-causal attention. 

\textbf{Off-policy distillation.}
This baseline evaluates on the teacher's trajectories instead of the student's. The frozen teacher
generates one trajectory per prompt; trace rows are built from the
committed positions of the teacher's trace, student and teacher are
evaluated on the teacher's pre-action states, and the full-vocabulary
reverse KL is minimized.

\textbf{TraceRL.}
For SDAR-4B, we evaluate the released TraDo-4B-Instruct, which was obtained from SDAR-4B-Chat with TraceRL and a diffusion value model on math and coding data. For SDAR-1.7B, no released checkpoint exists, so we train with the public implementation separately on the math and the code prompts: rollouts use the same dynamic decoding as our distillation runs, the update is the recipe's PPO-style objective with a KL penalty, rewards are the binary answer
match for math and the fraction of unit tests passed for code, and no
value model is used.

\textbf{TOPD.}
Our TOPD baseline is the trace-based on-policy distillation of
\citet{ren2026topd}, re-implemented in the code base of \ours{} so that
the two share the rollout, state reconstruction, and optimizer. Each
round samples one trajectory per prompt from the live student, builds
one row per denoising step with the committed positions as targets, and
applies the sampled-token estimator of \eqref{eq:k1}; the detached
teacher--student log-probability gap is clipped to $[-2, 2]$ before it
scales the gradient. Rows and positions are unweighted.

\textbf{\ours{}.}
\ours{} shares the rollout, row construction, and optimizer settings of
TOPD and changes only the objective to \eqref{eq:final}: every masked
position of every visited state is a target, the per-position loss is
the full-vocabulary reverse KL, and positions and rows carry the
hindsight weights $w_j$ and $w_b$ of Section~\ref{sec:method} at
their raw scale.

\subsection{Prompt Templates}
\label{app:prompts}
All SDAR runs use one prompt surface for student rollouts, teacher
supervision, SFT data generation, and evaluation. The templates below are
the ones in our training code; the evaluation code released with
TraceRL~\citep{wang2026revolutionizing} builds the same strings for math problems
and for stdio coding problems, and it prompts the coding benchmarks. The
teacher receives the same prompt as the student together with the
student's partially denoised response. Training uses only stdio coding
problems, since the PrimeIntellect snapshot contains no other kind. The
long instruction lines are wrapped here for display only; each template
keeps them on a single line.

\noindent\textbf{Math template.}
{\footnotesize
\begin{verbatim}
<|im_start|>user
{problem}
Please reason step by step, and put your final answer within
\boxed{}.<|im_end|>
<|im_start|>assistant
\end{verbatim}
}
\noindent\textbf{Code template (stdio).}
{\footnotesize
\begin{verbatim}
<|im_start|>user
This is the problem:
{problem}
You should put your code in ```python ```. Use input() to read input and
print() to produce output in your script. <|im_end|>
<|im_start|>assistant
\end{verbatim}
}

\subsection{Evaluation Protocol}
\label{app:protocol}
\textbf{Static and dynamic decoding.}
Both model families generate the response block by block. Within a block,
each denoising step proposes a token at every masked position, with a
confidence equal to the probability of the proposed token, and a decoding
rule chooses which proposals to commit; the others stay masked and are
proposed again at the next step. \emph{Static} decoding commits a fixed
number of positions per step, the $B/K$ with the highest confidence, so
every block takes exactly $K$ steps whatever its content.
\emph{Dynamic} decoding commits every position whose confidence exceeds
$\tau = 0.9$, and the single highest-confidence position when none does,
so the number of commits per step follows the model's confidence: a block
the model is sure about finishes in fewer steps, while an uncertain one
still takes up to $K$ steps. On SDAR we use the evaluation setting
released with TraDo and TraceRL~\citep{wang2026revolutionizing}: blocks
have $B = 4$ tokens and $K = 4$ steps, so static decoding commits one
token per step; sampling uses temperature $1.0$, top-$p = 1.0$, and a
2000-token limit; and a finished block enters the key-value cache before
the next block starts fully masked.

\textbf{Decoding on LLaDA.}
LLaDA-8B-Instruct attends bidirectionally over the whole sequence, so we
decode it with our reimplementation of the prefix-cache sampler of
Fast-dLLM~\citep{wu2025fastdllm}, following \citet{ren2026topd}. The
response starts fully masked and is filled in 32-token blocks. At the
first step of a block the model reads the prompt, the committed prefix,
the current block, and at most the next 128 masked positions; positions
beyond this horizon are left out of the input. The key-value cache of the
prompt and committed prefix is then kept, and the remaining steps of the
block recompute only the current block and its horizon. Each step
proposes, at every masked position of the block, the argmax of
Gumbel-perturbed logits at temperature $0.1$, and its confidence is the
probability of that token under the unscaled softmax. The commit rules
are those above with $K = 32$ steps per block, so static decoding again
commits one token per step; generation stops at 512 new tokens on math and 1,024 on
code. The temperature, block size, horizon, 512-token math budget, and
$\tau = 0.9$ follow the LLaDA evaluation of \citet{ren2026topd}. Since
they report no coding results, the 1,024-token code budget is the default
of the dLLM-RL LLaDA configuration~\citep{wang2026revolutionizing}.
Table~\ref{tab:inference} lists the inference settings of both
backbones.

\begin{table}[h]
  \centering
  \caption{Inference settings. SDAR settings follow
  \citet{wang2026revolutionizing}; LLaDA settings follow
  \citet{ren2026topd}, except the code budget, which is the dLLM-RL
  default. Static and dynamic decoding differ only
  in the commit rule and, on SDAR, in top-$k$. With $K = B$, static
  decoding commits one token per step, and dynamic decoding needs at most
  as many steps. On LLaDA, each proposal is the argmax of Gumbel-perturbed
  logits at the listed temperature, with no nucleus or top-$k$ truncation.
  \enquote{--}: not used.}
  \label{tab:inference}
  \small
  \begin{tabular}{l cc}
    \toprule
    \textbf{Setting} & \textbf{SDAR-4B / 1.7B-Chat} & \textbf{LLaDA-8B-Instruct} \\
    \midrule
    Max new tokens                          & 2000         & 512 (math), 1024 (code) \\
    Block size $B$ / steps per block $K$    & 4 / 4        & 32 / 32 \\
    Future horizon                          & --           & 128 \\
    Temperature                             & 1.0          & 0.1 \\
    Top-$p$                                 & 1.0          & -- \\
    Top-$k$ (static / dynamic)              & 1 / --       & -- / -- \\
    Commit threshold $\tau$ (dynamic)       & \multicolumn{2}{c}{0.9} \\
    \bottomrule
  \end{tabular}
\end{table}

\textbf{Evaluation library and grading.}
SDAR generation and the grading of both model families use dLLM-RL, the
evaluation code released with TraceRL~\citep{wang2026revolutionizing},
which runs SDAR through a vendored copy of the JetEngine inference
engine~\citep{cheng2026sdar}. We verified that it reproduces the
published TraDo-4B-Instruct numbers to within one point (MATH500 $76.2$
vs.\ $75.6$, GSM8K $91.3$ vs.\ $91.2$, static). MATH500, AIME2024, and
GSM8K answers are matched against the reference answer by dLLM-RL, first
as normalized strings and then for symbolic equivalence with SymPy. On
LiveCodeBench-v2 and LiveBench, a program
counts as correct only if it passes every hidden stdio test. The teacher row of Table~\ref{tab:main_results} is
quoted from \citet{ren2026topd}. The pilot study
(Section~\ref{sec:pilot}) instead decodes with our training-side JetEngine
under dynamic decoding and the training settings of
Section~\ref{sec:setup}, and grades math answers with OpenCompass.

\textbf{Checkpoint selection.}
\label{app:selection}
SDAR runs are evaluated every 5 rollout rounds on MATH500 (math) or
LiveCodeBench-v2 (code), and LLaDA runs every four rounds on MATH500
(math) or LiveBench (code). Under each decoding rule, the best round on
this selection benchmark is then evaluated on the remaining benchmarks.

\subsection{Wall-Clock Throughput}
\label{app:wallclock}
\begin{figure}[t]
  \centering
  \includegraphics[width=\linewidth]{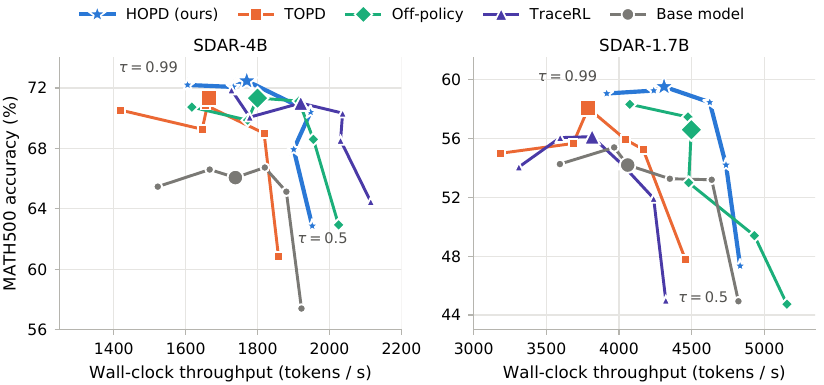}
  \caption{Accuracy against wall-clock generation throughput on MATH500
  (dynamic decoding), for the threshold sweep of Figure~\ref{fig:speed}.
  \textbf{Left}: SDAR-4B. \textbf{Right}: SDAR-1.7B.}
  \label{fig:wallclock}
\end{figure}

Figure~\ref{fig:wallclock} repeats the threshold sweep of
Figure~\ref{fig:speed} with wall-clock generation throughput on the
horizontal axis.  At every threshold the \ours{}
student generates $4$--$18\%$ more tokens per second than TOPD on SDAR-4B
and $8$--$23\%$ more on SDAR-1.7B.

\subsection{TOPD with Full-Vocabulary Reverse KL}
\label{app:topd_full_rkl}
The TOPD baseline in Table~\ref{tab:main_results} uses a sampled-token reverse-KL estimator, whereas \ours{} computes the reverse KL over the full vocabulary. To examine whether the advantage of \ours{} persists when TOPD uses the same per-position divergence, we additionally evaluate a full-vocabulary reverse-KL variant of TOPD. This variant retains supervision only at committed positions, with uniform position and row weights, but replaces the clipped sampled-token surrogate with full-vocabulary reverse KL. The remaining training settings, evaluation protocol, and checkpoint-selection rule are unchanged (Appendices~\ref{app:training} and~\ref{app:protocol}).

\begin{table}[t]
  \centering
  \caption{Comparison with full-vocabulary reverse-KL TOPD. Each entry reports static/dynamic scores; Avg.\ is the unweighted mean over the five benchmarks. Sampled-token TOPD and \ours{} results are repeated from Table~\ref{tab:main_results}. }
  \label{tab:topd_full_rkl}
  \small
  \setlength{\tabcolsep}{4pt}
  \resizebox{\linewidth}{!}{%
  \begin{tabular}{l cccccc}
    \toprule
    \textbf{Method} & MATH500 & AIME2024 & GSM8K & LCB-v2 & LiveBench & Avg. \\
    \midrule
    \multicolumn{7}{l}{\textbf{SDAR-1.7B-Chat}} \\
    TOPD (sampled-token)
    & 64.5/58.1 & 2.2/3.2 & 81.9/78.2 & 10.4/7.7 & 9.4/6.2 & 33.7/30.7 \\
    TOPD (full reverse KL)
    & 64.4/57.7 & 4.3/3.7 & 81.3/76.7 & 10.2/8.6 & 7.6/7.6 & 33.6/30.9 \\
    \rowcolor{gray!15}
    \ours{}
    & 66.8/59.5 & 3.7/5.0 & 83.6/78.8 & 11.1/8.9 & 10.9/9.9 & \textbf{35.2}/\textbf{32.4} \\
    \midrule
    \multicolumn{7}{l}{\textbf{SDAR-4B-Chat}} \\
    TOPD (sampled-token)
    & 74.9/71.3 & 11.2/7.2 & 92.0/90.0 & 21.9/18.6 & 19.8/17.4 & 43.9/40.9 \\
    TOPD (full reverse KL)
    & 74.8/70.9 & 7.8/9.2 & 91.6/90.8 & 22.0/20.2 & 22.7/19.8 & 43.8/42.2 \\
    \rowcolor{gray!15}
    \ours{}
    & 74.9/72.5 & 15.0/9.7 & 91.1/90.7 & 20.9/19.8 & 22.1/19.8 & \textbf{44.8}/\textbf{42.5} \\
    \bottomrule
  \end{tabular}}
\end{table}

Table~\ref{tab:topd_full_rkl} shows that full-vocabulary reverse KL has mixed effects on TOPD across tasks and decoding rules. In particular, it raises TOPD's five-benchmark average on SDAR-4B under dynamic decoding from $40.9$ to $42.2$, narrowing the gap to \ours{} ($42.5$). Nevertheless, \ours{} achieves a higher five-benchmark average than both TOPD variants at both model scales under both decoding rules. This comparison matches the per-position divergence but does not isolate the effect of supervising hesitations, since \ours{} also uses hindsight weights.

\section{Discussion and Ablation Details}
\label{app:ablation}

\subsection{Full-Attention Model Configuration}
\label{app:llada}
We use LLaDA-8B-Instruct~\citep{nie2026large} as the student and the
ESPO-trained LLaDA-8B-Instruct math and code models~\citep{ou2026espo}
as the corresponding teachers. The math teacher is merged from its
released LoRA. Table~\ref{tab:llada_hparams} outlines a four-GPU training
configuration based on the LLaDA recipe of \citet{ren2026topd}. For both
math and code, the effective batch size is 32 trace rows: 8 per GPU
$\times$ 4 GPUs $\times$ 1 gradient accumulation step.

\begin{table}[ht]
  \centering
  \caption{Experimental configuration for TOPD and \ours{}
  on LLaDA-8B-Instruct.}
  \label{tab:llada_hparams}
  \small
  \setlength{\tabcolsep}{5pt}
  \resizebox{\linewidth}{!}{%
  \begin{tabular}{l cc}
    \toprule
    \textbf{Setting} & \textbf{Math} & \textbf{Code} \\
    \midrule
    \multicolumn{3}{l}{\textit{Models and data}} \\
    Student & \multicolumn{2}{c}{LLaDA-8B-Instruct} \\
    Frozen teacher & ESPO-Math & ESPO-Code \\
    Training dataset & MATH (8,523 problems) & PrimeIntellect (5,954 problems) \\
    Max prompt length & 400 tokens & 512 tokens \\
    \midrule
    \multicolumn{3}{l}{\textit{Rollout (on-policy sampling)}} \\
    Inference engine & \multicolumn{2}{c}{Fast-dLLM} \\
    Prompts per round / responses per prompt & \multicolumn{2}{c}{32 / 1} \\
    Decoding rule & \multicolumn{2}{c}{low-confidence static} \\
    Block size / denoising steps per block & \multicolumn{2}{c}{32 / 32} \\
    Future horizon & \multicolumn{2}{c}{128 tokens} \\
    Sampling temperature & \multicolumn{2}{c}{0.8} \\
    Max response length & \multicolumn{2}{c}{512 tokens} \\
    \midrule
    \multicolumn{3}{l}{\textit{Optimization}} \\
    Optimizer & \multicolumn{2}{c}{AdamW, $\beta_1=0.9$, $\beta_2=0.999$, $\epsilon=10^{-8}$} \\
    Learning rate & \multicolumn{2}{c}{$2\times10^{-7}$} \\
    Weight decay & \multicolumn{2}{c}{0} \\
    Precision / gradient checkpointing & \multicolumn{2}{c}{bfloat16 / enabled} \\
    Parallelism & \multicolumn{2}{c}{4 GPUs, DeepSpeed ZeRO-2, CPU optimizer offload} \\
    Per-GPU batch size / gradient accumulation & \multicolumn{2}{c}{8 / 1} \\
    Effective batch size & \multicolumn{2}{c}{32 trace rows ($8\times4\times1=32$)} \\
    Gradient clipping (max norm) & \multicolumn{2}{c}{1.0} \\
    Rollout rounds / epochs per round & \multicolumn{2}{c}{30 / 1} \\
    Checkpoint cadence & \multicolumn{2}{c}{every 4 rounds} \\
    Seed & \multicolumn{2}{c}{1234} \\
    \midrule
    \textit{Distillation objective} & \multicolumn{2}{c}{top-$k$ reverse KL ($k=10$)}\\
    \bottomrule
  \end{tabular}}
\end{table}

\textbf{Training dynamics.}
Decoding, grading, and checkpoint selection follow
Appendix~\ref{app:protocol}; Table~\ref{tab:llada_rounds} gives the
selection-benchmark trajectories. On math, TOPD reaches its peak earlier and then decays,
whereas \ours{} starts lower, dips once around rounds 8--12, and ends higher. 

\begin{table}[h]
  \centering
  \caption{Selection-benchmark accuracy (static/dynamic) of the LLaDA-8B
  runs of Table~\ref{tab:full_attention} every fourth round: MATH500 for
  math runs, LiveBench for code runs. Bold: selected round per decoder.}
  \label{tab:llada_rounds}
  \small
  \setlength{\tabcolsep}{4pt}
  \resizebox{\linewidth}{!}{%
  \begin{tabular}{l ccccccc}
    \toprule
    \textbf{Run} & \textbf{4} & \textbf{8} & \textbf{12} & \textbf{16} & \textbf{20} & \textbf{24} & \textbf{30} \\
    \midrule
    TOPD, math    & 38.1/37.8 & 37.3/37.1 & \textbf{38.2}/37.6 & 38.0/37.5 & 37.5/\textbf{38.3} & 37.1/37.2 & 37.0/37.3 \\
    \ours{}, math & 37.6/37.1 & 37.0/36.7 & 37.6/36.0 & 38.5/37.0 & 38.2/37.5 & 37.7/37.2 & \textbf{39.0}/\textbf{38.1} \\
    \midrule
    TOPD, code    & \textbf{10.2}/\textbf{9.4} & 9.6/8.1 & 10.2/9.3 & 8.3/8.6 & 6.0/7.8 & 7.8/9.4 & 6.8/6.2 \\
    \ours{}, code & 9.4/5.5 & 10.9/9.4 & 11.2/9.4 & 9.9/6.2 & \textbf{13.0}/\textbf{10.7} & 9.6/9.9 & 10.9/7.0 \\
    \bottomrule
  \end{tabular}}
\end{table}

\subsection{Divergence Objectives}
\label{app:divergence}
The runs of Table~\ref{tab:divergence} keep the objective
\eqref{eq:final}, with the all-masked target set and both hindsight
weights, and the training configuration of Table~\ref{tab:hparams}; only
the per-position loss $\ell_{t,j}$ changes. Write
$p = \pi_\theta(\cdot \mid q, s_t, j)$ and
$r = \pi_{\mathrm{tea}}(\cdot \mid q, s_t, j)$ for the student and
teacher distributions at masked position $j$ of state $s_t$. The full
reverse KL is \eqref{eq:hes},
$\ell_{t,j} = D_{\mathrm{KL}}(p \,\|\, r)
  = \sum_{v} p(v) \log \frac{p(v)}{r(v)}$,
and the full forward KL swaps the arguments,
\begin{equation}
  \ell^{\mathrm{FKL}}_{t,j}
  = D_{\mathrm{KL}}(r \,\|\, p)
  = \sum_{v} r(v) \log \frac{r(v)}{p(v)}.
\end{equation}
JSD follows the generalized Jensen--Shannon divergence
of~\citet{agarwal2024policy}: with the mixture
$m_\beta = (1-\beta)\,p + \beta\, r$,
\begin{equation}
  \ell^{\mathrm{JSD}}_{t,j}
  = (1-\beta)\, D_{\mathrm{KL}}(p \,\|\, m_\beta)
  + \beta\, D_{\mathrm{KL}}(r \,\|\, m_\beta),
\end{equation}
and $\beta = 0.5$ gives the symmetric Jensen--Shannon divergence. These
three losses use the full vocabulary. The sampled-token loss reads only
the token $\hat{x}_j$ that the student proposed at position $j$ in state
$s_t$, whether or not it was committed, and applies the estimator of
\eqref{eq:k1} through the surrogate
\begin{equation}
  \ell^{\mathrm{ST}}_{t,j}
  = -\log p(\hat{x}_j)\;
    \operatorname{sg}\!\big[\log r(\hat{x}_j) - \log p(\hat{x}_j)\big],
\end{equation}
with the detached log-probability gap clipped to $[-2, 2]$ as in TOPD
(Appendix~\ref{app:baselines}).

\subsection{Hindsight Weights: Rounds and Code}
\label{app:weights}

\textbf{Setup.}
We compare uniform, position-only, block-only, and combined
weighting, keeping full-vocabulary
reverse KL fixed. Checkpoints are selected separately under
each decoding rule, using MATH500 for math and LiveCodeBench-v2
for code; the remaining benchmarks are evaluated at the selected
checkpoints. The combined-weight results reuse the \ours{} runs
in Table~\ref{tab:main_results}.

\textbf{Math results and training trajectories.}
Table~\ref{tab:weights_17b} extends the math ablation to SDAR-1.7B.
Combined weighting achieves the highest average under both
static and dynamic decoding ($51.4$ and $47.8$, respectively),
compared with $51.0$ and $46.7$ for uniform weighting.
Table~\ref{tab:weight_rounds} reports MATH500 accuracy every five
rounds for the two model scales in Tables~\ref{tab:weights}
and~\ref{tab:weights_17b}, with selected rounds in bold.
For the SDAR-1.7B position-only run, static MATH500 accuracy
declines from $62.2$ at round 20 to $47.3$ at round 30.

\textbf{Code results.}
Table~\ref{tab:weights_code} reports the corresponding code
ablations. Combined weighting achieves the highest average
on SDAR-4B under both static and dynamic decoding
($21.5$ and $19.8$, respectively), and on SDAR-1.7B under
static decoding ($11.0$).
Under dynamic decoding on SDAR-1.7B, position-only weighting
achieves the highest average ($9.9$ versus $9.4$ for combined
weighting). Thus, combined weighting performs well across
the evaluated settings, but the best weighting scheme can
depend on the task and decoding rule.

\begin{table}[t]
  \centering
  \caption{Hindsight weights on math for SDAR-1.7B-Chat, companion of
  Table~\ref{tab:weights}; same protocol. Bold: best per column.}
  \label{tab:weights_17b}
  \small
  \setlength{\tabcolsep}{4pt}
  \resizebox{\linewidth}{!}{%
  \begin{tabular}{l cccc cccc}
    \toprule
    & \multicolumn{4}{c}{\textbf{Static}}
    & \multicolumn{4}{c}{\textbf{Dynamic}} \\
    \cmidrule(lr){2-5}\cmidrule(lr){6-9}
    \textbf{Weights}
    & MATH500 & AIME2024 & GSM8K & Avg.
    & MATH500 & AIME2024 & GSM8K & Avg. \\
    \midrule
    uniform ($w_j = w_b = 1$)
    & 64.1 & \textbf{7.0} & 81.8 & 51.0
    & 58.4 & 4.2 & 77.6 & 46.7 \\
    $w_j$ only
    & 62.2 & 2.0 & 82.7 & 49.0
    & 57.1 & 4.0 & 77.6 & 46.2 \\
    $w_b$ only
    & 65.4 & 4.8 & 81.7 & 50.6
    & 58.6 & 3.8 & 77.0 & 46.5 \\
    \rowcolor{gray!15}
    $w_j$ and $w_b$ (\ours{})
    & \textbf{66.8} & 3.7 & \textbf{83.6} & \textbf{51.4}
    & \textbf{59.5} & \textbf{5.0} & \textbf{78.8} & \textbf{47.8} \\
    \bottomrule
  \end{tabular}}
\end{table}

\begin{table}[t]
  \centering
  \caption{MATH500 accuracy (static/dynamic, avg@3) every fifth round for
  the runs of Table~\ref{tab:weights}. Bold: round selected under each
  decoding rule (static/dynamic).}
  \label{tab:weight_rounds}
  \small
  \setlength{\tabcolsep}{4pt}
  \resizebox{\linewidth}{!}{%
  \begin{tabular}{l l cccccc}
    \toprule
    \textbf{Student} & \textbf{Weights} & \textbf{5} & \textbf{10} & \textbf{15} & \textbf{20} & \textbf{25} & \textbf{30} \\
    \midrule
    SDAR-4B   & uniform              & 74.4/70.6 & 74.8/\textbf{72.5} & \textbf{75.5}/71.2 & 73.1/70.7 & 75.4/71.7 & 74.6/71.4 \\
              & $w_j$ only           & 73.6/70.7 & \textbf{75.3}/70.1 & 75.1/70.3 & 74.0/71.0 & 74.3/70.9 & 73.9/\textbf{71.5} \\
              & $w_b$ only      & 73.1/71.5 & 72.8/71.8 & \textbf{75.5}/71.3 & 74.3/71.5 & 74.9/\textbf{72.3} & 74.3/71.8 \\
              & $w_j$ and $w_b$ & 73.3/71.3 & 74.6/70.5 & 74.5/71.6 & \textbf{74.9}/71.1 & 73.5/\textbf{72.5} & 73.9/71.4 \\
    \midrule
    SDAR-1.7B & uniform              & 59.2/56.1 & 61.8/56.7 & 63.0/56.5 & 61.8/57.4 & \textbf{64.1}/57.3 & 62.9/\textbf{58.4} \\
              & $w_j$ only           & 56.3/56.1 & 57.3/56.1 & 57.1/56.4 & \textbf{62.2}/\textbf{57.1} & 52.7/53.1 & 47.3/50.5 \\
              & $w_b$ only      & 63.7/58.1 & \textbf{65.4}/\textbf{58.6} & 62.7/58.1 & 60.5/57.5 & 54.2/54.9 & 65.1/57.7 \\
              & $w_j$ and $w_b$ & 62.5/58.4 & 59.7/57.3 & 58.0/58.0 & \textbf{66.8}/58.7 & 63.7/\textbf{59.5} & 63.0/58.1 \\
    \bottomrule
  \end{tabular}}
\end{table}

\begin{table}[h]
  \centering
  \caption{Hindsight weights on code, companion of Table~\ref{tab:weights}.
   Checkpoints selected on LiveCodeBench-v2 per
  decoding rule, LiveBench at the selected round.
  Bold: best per column.}
  \label{tab:weights_code}
  \small
  \setlength{\tabcolsep}{4pt}
  \begin{tabular}{l l ccc ccc}
    \toprule
    & & \multicolumn{3}{c}{\textbf{Static}} & \multicolumn{3}{c}{\textbf{Dynamic}} \\
    \cmidrule(lr){3-5}\cmidrule(lr){6-8}
    \textbf{Student} & \textbf{Weights} & LiveCodeBench-v2 & LiveBench & Avg. & LiveCodeBench-v2 & LiveBench & Avg. \\
    \midrule
    SDAR-4B   & uniform              & 21.2 & 21.5 & 21.4 & 19.5 & 18.0 & 18.8 \\
              & $w_j$ only           & \textbf{21.8} & 20.8 & 21.3 & \textbf{20.0} & 18.5 & 19.3 \\
              & $w_b$ only      & 20.6 & 19.3 & 20.0 & 19.5 & 19.3 & 19.4 \\
              & $w_j$ and $w_b$ & 20.9 & 22.1 & \textbf{21.5} & 19.8 & \textbf{19.8} & \textbf{19.8} \\
    \midrule
    SDAR-1.7B & uniform              & 10.3 & 9.4 & 9.9 & 8.7 & 9.1 & 8.9 \\
              & $w_j$ only           & 10.4 & 10.4 & 10.4 & \textbf{9.8} & \textbf{9.9} & \textbf{9.9} \\
              & $w_b$ only      & 10.0 & 8.6 & 9.3 & 8.3 & 8.3 & 8.3 \\
              & $w_j$ and $w_b$ & \textbf{11.1} & \textbf{10.9} & \textbf{11.0} & 8.9 & \textbf{9.9} & 9.4 \\
    \bottomrule
  \end{tabular}
\end{table}

\end{document}